\documentclass{article}
\usepackage[preprint]{colm2025_conference}

\usepackage{hyperref}
\usepackage{url}

\usepackage{lineno}

\definecolor{darkblue}{rgb}{0, 0, 0.5}
\hypersetup{colorlinks=true, citecolor=darkblue, linkcolor=darkblue, urlcolor=darkblue}

\usepackage{amsmath,amsfonts,bm}

\def\eqref#1{equation~\ref{#1}}

\def\1{\bm{1}}

\DeclareMathAlphabet{\mathsfit}{\encodingdefault}{\sfdefault}{m}{sl}
\SetMathAlphabet{\mathsfit}{bold}{\encodingdefault}{\sfdefault}{bx}{n}

\usepackage[utf8]{inputenc}
\usepackage[T1]{fontenc}
\usepackage{booktabs}
\usepackage{amsfonts}
\usepackage{nicefrac}
\usepackage{microtype}
\usepackage{graphicx}
\usepackage{natbib}
\usepackage{doi}
\usepackage{amssymb}
\usepackage{amsmath}
\usepackage{appendix}
\usepackage{algorithm}
\usepackage{algpseudocode}
\usepackage{import}
\usepackage{cleveref}
\usepackage{pifont}
\usepackage{mdframed}
\usepackage{xcolor}
\usepackage{tcolorbox}
\usepackage{tabularx}
\usepackage{longtable}
\usepackage[greek,english]{babel}

\tcbuselibrary{skins}

\newtcolorbox{importantbox}[1][Important Note]{
    enhanced,
    colback=blue!5,
    colframe=blue!75!black,
    arc=4mm,
    boxrule=1.5pt,
    title=#1,
    fonttitle=\bfseries,
    attach boxed title to top left={yshift=-3mm, xshift=5mm},
    coltitle=black,
    boxed title style={
        enhanced,
        colback=blue!40,
        arc=1mm
    },
    top=15pt,
    bottom=10pt
}

\newtcolorbox{optimismbox}[1][Value]{
    enhanced,
    colback=purple!5,
    colframe=purple!75!black,
    arc=4mm,
    boxrule=1.5pt,
    title=#1,
    fonttitle=\bfseries,
    attach boxed title to top left={yshift=-3mm, xshift=5mm},
    coltitle=black,
    boxed title style={
        enhanced,
        colback=purple!40,
        arc=1mm
    },
    top=15pt,
    bottom=10pt
}

\definecolor{empirical}{RGB}{230, 159, 0}
\definecolor{theoretical}{RGB}{0, 114, 178}

\title{A Mathematical Philosophy of Explanations in Mechanistic Interpretability \\
{\large The Strange Science: Part I.i}}

\author{Kola Ayonrinde
        \setcounter{footnote}{1}
        \thanks{
            Correspondence to:
            \texttt{koayon@gmail.com},
            \texttt{louis.yodj@gmail.com}
        } \\
        UK AI Security Institute
        \And
        {Louis Jaburi \textnormal{\textsuperscript{\textdagger}}}
}

\newcommand{\x}{\mathbf{x}}
\newcommand{\s}{\mathbf{s}}

\begin{document}

\maketitle

\begin{abstract}
    Mechanistic Interpretability aims to understand neural networks
    through causal explanations.
    We argue for the \emph{Explanatory View Hypothesis}:
    that Mechanistic Interpretability research is a principled approach to
    understanding models because neural networks contain implicit explanations
    which can be extracted and understood.
    We hence show that Explanatory Faithfulness,
    an assessment of how well an explanation fits a model, is well-defined.
    We propose a definition of Mechanistic Interpretability (MI) as the practice of producing
    \emph{Model-level}, \emph{Ontic}, \emph{Causal-Mechanistic}, and \emph{Falsifiable}
    explanations of neural networks,
    allowing us to distinguish MI from other interpretability
    paradigms and detail MI's inherent limits.
    We
    formulate the \emph{Principle of Explanatory Optimism},
    a conjecture which we argue is a necessary precondition for the
    success of Mechanistic Interpretability.

\end{abstract}

\section{Introduction}
\label{sec:intro}

ML artifacts are \emph{strange} objects.
ML researchers have produced models with a
wide range of cognitive capabilities that no human knows how to program
a machine to do,
from playing Go and poker at a superhuman level
\citep{schrittwieser2020alpha_zero, brown2019poker}
to folding proteins \citep{jumper2021alphafold}, and solving
advanced mathematical problems \citep{glazer2024frontiermath}\footnote{
  to deceiving humans in games
  \citep{golechha2025among_us_deception, meta2022diplomacy}
}.

However, we did not design these systems.
No human wrote the blueprint for how AI systems ought to perform a given task.
Instead, neural networks organically learn to solve problems
via gradient descent, given large quantities of data.
Neural networks aren't built; they're grown.

Because we don't design neural networks,
ML researchers typically do not know how their models perform a
given task.
Additionally, neural networks often solve problems in
unintuitive ways, relying on concepts that are not obvious to humans
\citep{widdicombe2018saliency_fingerprints,hosseini2018shape_bias, goodfellow2014og_adversarial_examples,ilyas2019adversarial_examples_features}.
This situation of relative ignorance about the processes that give rise to
a neural network's capabilities
leaves us with a scientific problem analogous to the natural sciences.
A physicist might observe some natural dynamical system, like the weather, and seek an explanation
allowing them to understand, predict, and possibly even steer the system.
Similarly, \emph{neural network interpretability} (henceforth just \emph{interpretability}) is the process of understanding artificial neural networks
using the scientific method.

In this way, we characterise Interpretability as \emph{The Strange Science}:
Interpretability is the science of understanding artificial neural phenomena,
just as the natural sciences seek to understand natural phenomena.
Interpretability researchers study formal systems using empirical methods ---
making observations, generating conjectures, and refuting those conjectures ---
to understand complex neural systems.
Since interpretability is analogous to the natural sciences,
a Philosophy of Interpretability
should be inspired by our best understanding of the philosophy of science.

Recent works like
\citet{mech_interp_review_2024}, \citet{sharkey2025openproblemsmechanisticinterpretability},
and \citet{geiger2023causal}
have explored the methods and assumptions of
Mechanistic Interpretability (MI).
Other works have explored the philosophically relevant components of MI
\citep{milliere_philosophy_interventionist_methods, harding2023representations_nlp, kastner2024interp_as_xai_philosophy}.
In this work, we foreground the philosophical role of \emph{explanation} in \emph{Mechanistic Interpretability} specifically,
and how this differs from previous interpretability paradigms.\footnote{
    \citet{lipton2018mythos_classical_interp,gilpin2018explaining_explanations,Fleisher_2022_explainable_ai,doshi2017rigorous_interp,leavitt2020falsifiableinterpretabilityresearch,erasmus2021interpretability}
    provide an
    overview of classical (pre-Mechanistic) Interpretability.
}
In particular, inspired by the Information-Theoretic perspective,
we can understand explanations in terms of their compressive power
and ability to communicate \emph{understanding which generalises}.

Understanding neural networks can provide affordances for interventions which
are important for AI Safety, AI Ethics, and AI Cognitive Science
\citep{bengio2025ai_safety_report, anwar2024foundational, chalmers2025propositional,
olah2020zoom_in_circuits,Amodei_2025_urgency_interpretability}.
Such understanding also allows us to improve the performance of neural networks
and debug their failings
\citep{lindsay_bau_interp, sharkey2025openproblemsmechanisticinterpretability,Amodei_2025_urgency_interpretability}.

\paragraph{Contributions.} Our contributions are as follows:
\begin{itemize}
\item
  Firstly, we show that producing compressive explanations frames a potential solution
  to the Interpretability Problem.
  We hence define \emph{explanatorily faithfulness} as the goal of Mechanistic Interpretability.
\item
  Secondly, we provide a technical definition of MI and leverage this
  definition to highlight both the possibilities and limitations of MI.
\item
  Thirdly, we formulate the \emph{Principle of Explanatory Optimism},
  a conjecture at the heart of MI which states that the
  algorithmic structure of generalising neural networks
  is human-understandable.
  We show that without the Principle of Explanatory Optimism
  the project of MI is intractable.
\end{itemize}

\paragraph{Series Structure}
This paper is the first in a series titled
\emph{The Strange Science of Mechanistic Interpretability},
concerning the Philosophy of Mechanistic Interpretability. \\
See later papers in this series for an evaluation of methods in MI
through the lens of Explanatory Virtues
\citep{ayonrinde_2025_expl_virtues_ss1_2} (Part I.ii).
See also \citet{ayonrinde2025bidirectional_interp} (Part I.iii)
which proposes methods to empower humans by teaching humans
Machine Concepts.

\section{The Strange Science}
\label{sec:strange_science}
Is Interpretability a natural science or a formal science?
\emph{Natural sciences} like physics, biology, and earth sciences
seek to explain physical phenomena,
creating hypotheses and running experiments.
\emph{Formal sciences} like algebra, decision theory, linguistics, and music theory
are concerned with abstract systems and structures,
using deductive methods to construct proofs from axioms.
The defining strange property of neural network interpretability
is that it is a natural science whose objects of study are both
\emph{artificial} (constructed by humans rather than naturally occurring)
and \emph{formal} (inherently abstract rather than physical systems).

Neural networks are mathematical objects; they are deterministic
functions from the input embedding space to the output embedding space.
Yet, the strange science of interpretability is not a formal science.
Interpretability researchers are not primarily interested in proving theorems about neural networks,
but in understanding the empirical properties of neural networks.
They seek to understand ``why'' questions about generalisation:
\emph{why does the network generalise in this specific way?}
or \emph{what are the reusable representations that a neural network is using?}
and so on.

We a priori know the model's weights, architecture and formal specification
and we can compute the output behaviour corresponding to any given input.
Where other sciences might be limited by the precision
of their measuring tools or the fidelity of observations,
in Interpretability our observations are exactly precise
and our experiments perfectly reproducible.
Furthermore, we can
intervene on the network at any point and any time to
arbitrarily high precision.
However, despite this formal knowledge and potential for intervention,
understanding neural networks remains elusive.
Here we find a peculiar reversal of the sciences:
in natural science, we pursue mathematical formalism to describe empirically observed phenomena;
yet in interpretability, we pursue empirical methods to understand formalism.

To make progress in understanding ML models,
we approach them as naturalists studying a complex system.
We can see neural networks as an exemplar of the pitfalls of reductionism:
we know all the material and formal causes of the network,
and we understand each individual part;
however, we do not understand the system.
We have complete access to all formal properties of the system,
yet our \emph{scientific knowledge} of the system is incomplete.
The \emph{strange} paradox of interpretability is thus:
we have a complete understanding of neural networks at the
base, formal, implementation level up to arbitrary precision.\footnote{
    This would be comparable to a physicist having perfect knowledge
    of the fundamental particles and forces of the universe,
    being able to measure and manipulate each atom at will and
    see subatomic particles with the naked eye.
    Or a biologist knowing the precise reactions occurring
    in every cell in the body and knowing the structure and shape
    of every relevant molecule
    (though see also \citet{Jonas_neuro_microprocessor}).
}
However, it's not immediately obvious how this low-level
knowledge translates into high-level understanding
and the ability to predict and control
the complex system's behaviour.
We might say that the properties of a neural network \emph{supervene} on low-level mathematical facts
about the system.
That is, the neural network is entirely defined by its low-level facts,
but yet there are \emph{emergent} mental phenomena that are
not apparent purely by analysing the low-level facts.

To understand a neural network, we would like to understand the
relevant variables in model's computation.
These variables, which we might call \emph{features}, are
generally not neurons or network parameters;
they are instead unseen entities that we must posit and discover
like subatomic particles in physics.\footnote{
    The Sparse Autoencoder paradigm has a specific
    linearly accessible interpretation of these features;
    in general we make no such commitment to any particular
    instantiation of how features are represented in the network.
}
Though we have perfect formal knowledge of the system,
we are nonetheless explaining what we see
(the network's behaviour) in terms of what we cannot immediately
see (the features).
In interpretability, as in other natural sciences,
understanding of the seen is
revealed through the unseen
\citep{deutsch2011beginning_of_infinity, marion_visible_revealed, Girard1987_hidden_since_foundation, Aquinas1273_summa}.

\paragraph{Paper Structure.}
The rest of this paper is organised as follows:

\begin{itemize}
    \item
        In \Cref{sec:explanations_and_interp}, we provide an exposition
        of the \emph{Explanatory View} of
        neural networks: a model's internal structures admit explanations of
        model behaviour.
        We argue that the Explanatory View provides additional justification
        for why MI researchers can productively seek
        Causal-Mechanistic explanations of ML models.
    \item
        \Cref{sec:mech_interp_def} provides a technical definition
        of Mechanistic Interpretability as seeking model explanations
        that are \emph{Model-level},
        \emph{Ontic}, \emph{Causal-Mechanistic}, and \emph{Falsifiable}.
    \item
        From this definition, we analyse the inherent limits of
        Mechanistic Interpretability in \Cref{sec:limits} and discuss
        implications of the Explanatory View in \Cref{sec:discussion}.
    \item
        We conclude in \Cref{sec:explanatory_optimism} by articulating
        the implicit conjecture at the heart of interpretability,
        which we call \emph{The Principle of Explanatory Optimism} (EO).
        EO states that the generalising algorithms
        learned by neural networks are human-understandable.
\end{itemize}

\section{Explanations and Interpretability}
\label{sec:explanations_and_interp}
Much of the human experience, both social and personal, is made up of explanations.
We explain why vegetables are good to our children,
why product A is likely to sell more units than product B to our boss,
why to pursue a given research topic to ourselves, and so on.
But what do we mean by the term `explanation' in science?

\subsection{Scientific Explanation}
\label{sec:what_is_an_explanation}

The epistemic aim of science is to understand phenomena
by way of explaining these phenomena
\citep{DeRegt2017_scientific_understanding}.
A scientific explanation, then, is an answer to a ``why'' question
\citep{lipton2001good_of_explanations}.
The fundamental question that explanations answer is
``why did the phenomenon occur?'' \citep{hempel_explanation}.
We can view explanations as a solution to a problem:
there's a gap between our current best theory
and the phenomena that we would like to explain.
Good explanations close this gap.
With a good explanation, we can say that the phenomenon
was indeed expected --- and crucially --- here's why.
Explanations are vehicles for understanding;
someone understands a phenomenon when they grasp an
accurate explanation of the phenomenon
\citep{strevens2013no_understanding_wo_explanation,Khalifa2013_explanation_for_understanding}.

\paragraph{Understanding and Compression.}
\citet{wilkenfeld2019understanding_compression} describes the close relationship
between understanding and compression.\footnote{
    See also \citet{chaitin2002intelligibility, li2008kolmogorov, mackay2003information, solomonoff_induction,hutter2023aixi}.
}
Given a series of observations which characterise a phenomenon,
we understand the phenomenon
if we have an explanation that compresses the data into a more concise form
such that we could reproduce the data from the explanation or
use the explanation to predict future data.
Good explanations exploit regularities in the data for compression.
A series of observations is incomprehensible if it cannot be compressed,
that is, if it contains no regularities to exploit and is purely random \citep{li2008kolmogorov}.
Explanations are not \emph{merely} compressions however.
It is not obvious that a compressed zip file engenders more understanding
than the original data file in general.
Explanations are compressions of a particular kind:
those compressions that facilitate the understanding of phenomena
\citep{ayonrinde2024_mdl_saes}.

\subsection{From Induction To Explanation}
\label{sec:induction_to_explanation}

\citet{andrews2023theory_free} details a classical view of machine learning as a process of induction:
``ML models use evidence, or training data, to form predictions or classifications,
which generalise what they have learned from their training set to unseen instances (i.e., novel data).
The field of ML strives to automate inductive inference.''
When we view ML models in this behaviourist fashion as black boxes
with only inputs and outputs,
we may think of them as providing predictions without explanations.
Such \emph{explanationless predictions} are of the same variety as a prophecy from an oracle.

Suppose that we are to think of an ML model qua oracle as providing reasons to believe
some proposition $p$.
Since oracles do not provide explanations in terms
of the prediction's \emph{content} (subject matter),
we may only believe such an explanationless oracle-style claim if we have
extrinsic reasons to believe the model's prediction
(for example, that the model has been generally correct before or
that we have some knowledge of the model's training or similar).
For MI researchers, however, believing a proposition $p$ from an ML model
should be based on the \emph{content} of model explanations rather than merely on extrinsic reasons
like the model's track record (see \Cref{sec:appendix_oracles}).

Mechanistic Interpretability researchers do not view neural networks
as incomprehensible inductive black boxes.
In this paper, we offer a philosophical substantiation of Mechanistic Interpretability
as practised by scientists.
We will argue that MI researchers take an alternative Deutschian \citep{deutsch2011beginning_of_infinity}
\emph{Explanatory View} of neural networks, which we may contrast
with the classical inductive perspective expressed by \citet[inter alia]{andrews2023theory_free}.

Where the classical view understands neural networks as black boxes that inductively infer from data,
the Explanatory View would have us consider knowledge of the internal mechanisms
of the model as necessary for understanding the model's behaviour.
Here the internal mechanisms themselves can be seen as implicit explanations of the model's behaviour -
the reason that a model makes a prediction is contained within its internal mechanisms.
This is a white-box (cognitivist) view of neural networks.
In this way, we can view models as \emph{proto-explainers} rather than merely predictors.

\subsubsection{Generalisation: What We're Explaining When We're Explaining Neural Networks}
\label{sec:what_we're_explaining}

When we are interested in explaining neural networks, what we would like to explain
is how, and in which ways, they \emph{generalise}.
By generalise, we mean that the model can leverage regularities \& structure
in the training data to solve some task with respect to unseen data.
As models learn to generalise,
\emph{internal structure} forms within the model.\footnote{
    Here we are interested in explanations
    of neural networks as objects of scientific and philosophical interest.
    Much previous work has been interested in explanations of the results of
    neural networks as it pertains to some task (e.g., medical diagnosis).
    Here the application domain is of secondary interest
    and we are examining explanations \emph{of neural networks themselves}.
}

A system contains structure to the system's generating process can be expressed
more concisely (i.e., in fewer bits) than the observations of the system.
That is, structure is \emph{compressibility}.
Systems that follow general principles contain structure and regularities such that they are
(at least in principle) more predictable than structureless systems
like random noise generators.
For example, consider an idealised pendulum's position over time.
The pendulum's position follows a predictable pattern which can be
expressed as a mathematical function and hence we only need to store the
equation and the initial conditions to reproduce the observations of the
pendulum's position over time.

In this sense, the natural world contains structural patterns ---
there are natural laws which allow us to compress and understand the world.
And the training data for ML models, which is sampled from the world,
inherits such structure from the natural world.
ML models generalise to the extent that they can learn or approximate the
structure in the world through the training data.\footnote{
    Models trained on randomised
    data may memorise but never learn to generalise
    \citep{zhang2017understanding_dl_generalisation, lehalleur2025slt_position,
    lin2017whydeeplearning,deletang2024lm_compression}.
}
A good explanation should expose the model's internal learned structures.
We define \emph{ur-explanations}\footnote{
    Here we use the \emph{ur-} prefix to indicate primacy or origin as in the German language.
}
as the idealised explanations of model behaviour on an input distribution,
given in terms of its learned internal structures.\footnote{
    Considering the model internals as ur-explanations does not necessarily
    mean that all such ur-explanations will be interesting explanations of generalisation per se.
    Models may have memorised certain answers or resort to bags of faulty heuristics in some cases.
    However, in at least some cases, we have evidence
    of models learning genuine algorithms which represent explanatory knowledge within the network
    \citep{wang2024neurips_grokking, nanda2023grokking, wu2024groups_unified_compact_proofs}.
}

The \emph{ur-explanations} of a neural network can be seen as \emph{internal computations} over
\emph{learned representations} that compute the output from the input.
The network learns these representations during training in a process of
automated Conceptual Engineering (see \Cref{sec:conceptual_engineering}).
These network computations, outputs, and intermediate activations together
constitute not only a prediction of some answer,
but also an explanation of the process by which the model came to such a result.

\subsubsection{Explanatory Faithfulness}
\label{sec:explanatory_faithfulness}

The Explanatory View takes seriously the idea that there is structure in the
model to be interpreted (see also \Cref{sec:three_varieties}).
Under this view, there is a target to the interpretability program:
we are not merely looking for explanations that appear to correlate with model behaviour,
we are looking to extract the internal explanations from neural networks.
Explanations can be more than confabulatory just-so stories that provide the illusion
of understanding the model's behaviour.

Under the Explanatory View, we can now define \emph{explanatory faithfulness}.
An explanation is explanatorily faithful to the model to the extent that it
matches the model's ur-explanation.\footnote{
    We argue for the uniqueness of the ur-explanation in \Cref{sec:nma}.
}
Interpretability researchers would like to say that their explanations
are faithful to the model and (approximately) describe the same
algorithmic mechanisms that the model uses.
Note the difference here between our notion of \emph{explanatory faithfulness}
and \emph{(behavioural) faithfulness} (e.g., from \citet{wang_ioi}):
\emph{behavioural faithfulness} says that the explanation and model produce the
same outputs; \emph{explanatory faithfulness} says that the step-by-step explanation
matches the model's internal mechanisms, not just the input-output behaviour.
Note that defining explanatory faithfulness is not possible
under the classical view of Machine Learning without ur-explanations.
Under the classical view, statements about circuit equivalence
can only be understood as statements about behavioural statistics
\citep{shi2024hypothesis_testing_circuits}.

\begin{importantbox}[Explanatory Faithfulness]

    An explanation $E$ is \textbf{explanatorily faithful} to a model $M$ over some data distribution $D$,
    to the extent that  intermediate activations $\s_i$ at each layer $i$ that
    are given by the algorithmic explanation $E$ closely match
    the intermediate activations $\x_i$ of the model $M$ for input data in $D$.

\end{importantbox}

\subsection{Neural Networks Perform Computations over Representations}
\label{sec:computations_over_representations}

We described the ur-explanation of a model
(the idealised explanation of model behaviour)
in terms of \emph{Computations} over \emph{Representations}.
We now provide more details on what is meant by each of these terms.

\paragraph{Computation.}
\citet{marrs_levels} describes 3 levels of analysis for understanding a machine
carrying out an information-processing task
\citep{McClamrock1990_marr_revisited,Angelou_2025_Three_levels}:

\begin{enumerate}
    \item \textbf{Computational Level:}
        What is the goal of the computation, why is it appropriate, and what is the logic of the strategy by which it can be carried out?
    \item \textbf{Algorithmic/Representational Level:}
        What is the algorithm being used to perform the computation?
        How can this computational theory be implemented?
        In particular, what is the representation for the input and output, and what is the algorithm for the transformation?
    \item \textbf{Implementation Level:}
        What is the physical implementation of the algorithm?
        How can the representation and algorithm be realized physically
        on some computational substrate?
\end{enumerate}

For neural networks, the \emph{Implementation} Level of Analysis
corresponds to matrix multiplications with the model weights and how these
are implemented on the substrate of hardware computational accelerators
\citep{Angelou_2025_Three_levels}.
We know essentially all there is to know formally about the
Implementation level.

When we speak of a neural network carrying out computations,
we are referring to the \emph{Algorithmic} and
\emph{Computational} Levels of Analysis.
We would like to have useful compressive causal explanations
at the Algorithmic and Computational levels which
are detailed enough to be ``runnable''
\citep{cao2024explanations_part1}.\footnote{
    \Cref{sec:straightforward_explanation} details
    an Implementation-level explanation of a neural network.
    By ``runnable'' here we mean that the explanation that we provide
    should be pseudocode that we could imagine formalising
    such that it would compile and run on a computer.
}
Explanatory Faithfulness is an Algorithmic level property:
the stages of the explanation should match
the model layers and be ``locatable''.
It is not sufficient for Explanatory Faithfulness for the outputs
to agree if algorithms producing the outputs differs.

\paragraph{Representation.}
A pattern of neural activations
is a \emph{representation} when it represents \emph{something},
that is it has some appropriate correspondence with features
of the input data (and hence the external world).
Representations are representations \emph{of}
a feature.\footnote{
    In other words, representations have \emph{intentionality}.
    Note that we use the word intentionality here
    in the philosophical sense of ``aboutness''.
    This usage of the term is not to be confused with the psychological sense of
    ``intention'' (as in I intend to get the next train home)
    or any claims about some conscious relationship to representations.
}
We paraphrase \citet{harding2023representations_nlp}'s three criteria for
activations to qualify as representations below.

Consider a pattern of activations $h(\x)$ for $\x \in X$, where $X$ is the domain
of a model (e.g. natural language).
Then $h(\x)$ \emph{represents} a property $Z$ if the following three criteria hold:

\begin{itemize}
    \item \textbf{Information:}
        The activations $h(\x)$ correlate with the property $Z$.
        More formally, the random variable $h(\x)$ has sufficiently high Shannon
        Mutual Information with the property $Z$, $I(h(\x); Z)$, such that
        we could train a successful probe $g_z : h(\x) \to \mathcal{P}(Z)$.
        Intuitively, Information says \emph{representations are \textbf{Causal Results} of features
        contained within the input $\x$}.
    \item \textbf{Use:}
        The model uses the information in activations $h(\x)$ about $Z$
        to perform its task.
        That is to say that if we were to remove the relevant information from
        the activations through a causal intervention,
        the model's performance on the relevant downstream tasks would decrease.
        Intuitively, Use says \emph{representations are \textbf{Causes} of the model's behaviour}.
    \item \textbf{Misrepresentation:}
        It should be possible for the activation vector $h(\x)$ to misrepresent $Z$.
        Suppose that we have activations $h(\x)$ which do not contain useful information about $Z$
        and $h(\s)$ which does contain information about the property $Z$.
        Then we say that $Z$ is misrepresentable if we can perform an intervention which patches the information $h(\s)$ into $h(\x)$,
        and predictably increase the likelihood of our model mistaking our input $\x$ for having property $Z$.
        Intuitively, Misrepresentation says \emph{representations can be \textbf{causally intervened}
        on}.\footnote{in the sense of Causal Abstractions Theory
        \citep{geiger2023causal,pearl2009causality,Beckers_halpern_2019_Abstracting_Causal_Models}}
        To be able to represent, you must be able to misrepresent.
\end{itemize}

A pattern of neural activations that satisfies the Information, Use and
Misrepresentation criteria can be called a \emph{representation}.

\subsection{The Goal of Interpretability}

ML methods are often applied to problems in other fields
like predicting weather patterns, classifying legal cases and allocating scarce resources.
Interpretability, then, can be viewed as applying ML methods and analysis to an epistemic problem:
``how does a neural network perform computations over representations
to produce useful answers to queries?''

Interpretability researchers don't want to only know what a neural network predicts.
We would also like to understand the structures, features, regularities, and knowledge
which cause the neural network to make such and such a prediction.
We would like to extract \emph{explanatory knowledge} from neural networks,
\emph{uncovering} the ur-explanations that are always-already present
within a trained, generalising model.
Most ML researchers are in the prediction business.
Interpretability researchers, however, are in the explanation business.

The Explanatory View treats neural networks
as containing explanations rather than as being purely behaviourist oracles,
moving from black-box induction to white-box computations.
The Explanatory View is the first step in understanding ML models
not in terms of prediction but in terms of \emph{explanation}.

\section{Demarcating Mechanistic Interpretability}
\label{sec:mech_interp_def}
There has been much discussion about what makes some interpretability research `Mechanistic'
rather than another form of interpretability \citep{saphra2024mechanistic, chalmers2025propositional}.
\citet{Gieryn1999} describes the problem of demarcating
where one science starts and another begins —`boundary-work'— analogously to the
Demarcation Problem between Science and Pseudo-Science \citep{Laudan1983, popper_scientific_discovery}.
The definition of a given science can be seen as a grab-bag of associations like
Wittgensteinian language games \citep{Wittgenstein1953}.
Another way to define a science is as (social) culture \citep{Bruno1987}.
Under this view ``Science is what scientists do'' \citep{Bridgman1980}.

To formalise Mechanistic Interpretability, we instead provide a technical definition
that focuses on the goal of Mechanistic Interpretability compared
to other adjacent disciplines \citep{olah2020zoom_in_circuits, saphra2024mechanistic}.
We define Mechanistic Interpretability as the study of
\emph{Model-level, Ontic, Causal-Mechanistic, and Falsifiable Explanations} of Neural Networks.
This definition clearly delineates Mechanistic Interpretability from
other paradigms like Concept-Based Interpretability.\footnote{
  This delineation is useful for researchers but we do not
  intend to imply that non-Mechanistic Interpretability is not useful,
  see \Cref{sec:appendix_mi_value}.
}
We can understand these properties of explanations by way of
contrast with other forms of explanation.\footnote{
  We provide intuitive examples of explanations with
  these properties in \Cref{sec:examples}.
  We further compare Mechanistic Interpretability with previous
  interpretability paradigms in \Cref{sec:three_varieties}.
}

\paragraph{Model-level Explanations.}
An autoregressive language \emph{model} is a neural network that returns a probability
distribution over possible next tokens when conditioned on some input tokens
(i.e., textual prompt).
A language model \emph{system} (LM system) is a software object
that contains a language model as part of the control flow.
The LM system leverages the language model to produce some useful output,
like a text completion or an image,
rather than a single next-token probability distribution.
An LM system may be as simple as augmenting a language model with a sampling method or greedy decoding.
More complex LM systems may use meta-decoding strategies, tool use, automated prompting,
multiple language models, and more
\citep{arditi_2024_systems_interp, compound-ai-blog, khattab_dspy,
guo2024multi_agents, dafoe2020cooperative_ai, welleck_meta_decoding_tutorial}.

Capability evaluations are typically system-level evaluations.
Model performance depends substantially
on prompting strategies like Chain of Thought reasoning \citep{wei2022chain_of_thought}
and tool use \citep{schick2023toolformer}.
Systems-level explanations might seek to explain whole system performance by, for example,
reading a model's Chain of Thought \citep{perez2023reading_cot}.
Conversely, Model-level explanations seek to understand the neural
network part of the system in isolation
to explain why the output distribution is as it is.

\paragraph{Ontic Explanations.}
Ontic explanations consist of real, physical entities \citep{causal-realism-salmon}.
We may contrast Ontic explanations with epistemic explanations
which focus on making phenomena understandable
or predictable to the interpreter,
potentially using idealizations, models, or abstractions
that may not directly correspond to reality.
Non-ontic, ``epistemic'' explanations may give useful rules of thumb for prediction
or intuition but may not be well supported by reality.\footnote{
    Note that scientific non-realists may only produce epistemic explanations,
    as they may not believe that the entities referred to in scientific theories actually exist
    in reality. See also \Cref{sec:nma}.
}
\citet{varma2023circuit_efficiency_grokking}'s work on circuit efficiency can be seen as giving non-ontic explanations,
through the hypothesised efficiency metric.

\paragraph{Causal-Mechanistic Explanations.}
\label{paragraph:causal-mechanistic}

Causal-Mechanistic Explanations
\citep{woodward2003-causal_explanation, salmon1989four_decades_scientific_explanation,Lewis1986_causal_explanation}
identify the causal processes that produce phenomena
rather than just describing statistical correlations or general laws.
Here, we are interested in the relevant components of a system, how they are organised,
and how they interact to produce phenomena.
Causal-Mechanistic theorists refer to these explanations as “explaining why by explaining how”
\citep{Bechtel2005_mechanistic_explanation}:
explaining why a phenomenon occurred involves identifying the
underlying mechanisms that give rise to observed phenomena.
Causal-Mechanistic Explanations go step by step to explain the end-to-end process:
they provide a continuous causal chain from cause to effect,
without any unexplained gaps.
Causal-Mechanistic Explanations explain the end-to-end process
\citep{causal-realism-salmon,lipton_causal_model}.\footnote{
    Several works within Machine Learning that also provide a good introduction to causal modelling include
    \citet{jintutorial_causality, scholkopf2021causal_representation_learning, liu2024llms_causal_inference, pearl2009causality}.
}

We can contrast Causal-Mechanistic Explanations with:
\begin{itemize}
  \item
    \textbf{Statistically Relevant Explanations} \citep{salmon_Statistical_explanation, salmon1989four_decades_scientific_explanation}.
    X explains Y if and only if
    $P(Y|X) \neq P(Y)$ — that is,
    if and only if the conditional probability of Y given X differs from the probability of Y.
    For example, we might explain ice cream sales by high correlation with
    temperature.\footnote{
        or unhelpfully, our explanation could show correlation with the number of shark attacks
    }
  \item
    \textbf{Telic Explanations} \citep{Sosa2021_telic_explanation}.
    We explain a phenomenon by reference to its purpose, aims, or function
    rather than in terms of a causal chain of events.
    For example, we might explain the heart as being for the purpose of
    transmitting and pumping blood.
  \item
    \textbf{Nomological Explanations}
    \citep{myers2012cognitive_styles_nomological, Scheibe2002-rationalism_empiricism}.
    We explain a phenomenon by reference to general laws or principles
    rather than in terms of a causal chain of events.
    For example, linguistic theory might appeal to universal grammar
    ``laws'' to explain the structure of human languages.
\end{itemize}

\begin{importantbox}[Technical Definition of Mechanistic Interpretability]

    Interpretability explanations are \textbf{valid} as Mechanistic Interpretability explanations
    if they are
    \textbf{Model-level}, \textbf{Ontic}, \textbf{Causal-Mechanistic} and \textbf{Falsifiable}.

\end{importantbox}

Causal-Mechanistic and Ontic explanations are necessary to the empirical
practice of Mechanistic Interpretability amongst active researchers
\citep{mech_interp_review_2024,sharkey2025openproblemsmechanisticinterpretability}.
Though it is feasible to imagine a causal-mechanistic approach to understand
system-level behaviours, it is a historically contingent fact that the field
has coalesced around methods for model-level explanations \citep{saphra2024mechanistic}.\footnote{
    Since system-level explanations are of practical and
    academic interest,
    there is currently an opportunity for a new field to emerge focusing on system-level explanations,
    possibly building on the work of the mechanistic interpretability community.

    Perhaps the nascent field of LLM-ology might be a candidate to fill this gap \citep{Trott_2023_llmology}.
    Note that many Benchmarks and Evaluations researchers could be seen as working in this space already.
    Chain of Thought interpretability \citep{perez2023reading_cot} is another example of system-level explanation.
}

\section{The Limits of Mechanistic Interpretability}
\label{sec:limits}

We have analysed the type of explanations that
Mechanistic Interpretability researchers seek, namely those which are
Model-level, Ontic, Causal-Mechanistic and Falsifiable
Explanations of a model's internal mechanisms.
We now turn our attention to the extent of the limits
and challenges of such explanations.

\subsection{Value-Ladenness \& Theory-Ladenness of Explanations}
\label{sec:value_ladenness}

We would like explanations that are accurate and human-understandable
compressed representations of observations.
With this goal in mind, the best explanation of a phenomenon is interpreter-relative
in the following two senses.
\emph{Firstly}, the ideal explanation is relative to the interpreter's
initial set of concepts, their priors and what types
of explanation are easy for them to understand.
In this sense, explanations are \textbf{Theory-Laden}.
\emph{Secondly}, the ideal explanation depends on what the interpreter would like to \emph{do}
with such an explanation. The interpreter may be satisfied with a
different level of granularity of explanation depending on whether
they are seeking an explanation of a model's behaviour to be able
to make crude interventions, or to make guarantees about model performance,
or for scientific curiosity.
In other words, the ideal explanation is also relative to the
interpreter's \emph{values}; explanations are \textbf{Value-Laden}.

\subsubsection{Value-Ladenness of Explanations}

\citet{weber1949objectivity_value_free} argued for the
Value-Free Ideal in the sciences, the principle that scientists should be value neutral.
We can articulate the Value-Free Ideal as
``\emph{Scientists should strive to minimize the influence of contextual values
on scientific reasoning, e.g., in gathering evidence and assessing/accepting scientific theories}''
\citep{sep-scientific-objectivity}.

For the normative statement of the Value-Free Ideal to be considered a reasonable ideal,
it must be attainable (at least to some degree).
That is, ought implies can.
So we may first analyse whether value-freeness is possible which can be expressed in the
Value-Neutrality Thesis as follows:
``\emph{Scientists can—at least in principle—gather evidence and
assess/accept theories without making contextual value judgments}''
\citep{sep-scientific-objectivity}.

In science generally, and interpretability research particularly, it is difficult to
hold Value-Neutrality. Hence the Value-Free Ideal seems to be unattainable and
likely undesirable as a goal \citep{Douglas2009value_free_ideal}.
The choice of methods and which results are particularly interesting for researchers
has a close dependence on what researchers might hope to achieve.
Many researchers in Mechanistic Interpretability are interested in the applications
to AI Safety, AI Ethics, AI Cognitive Science and AI Governance
\citep{bengio2025ai_safety_report, anwar2024foundational, olah2020zoom_in_circuits},
all of which affect researchers' contextual value judgements.
Evidential standards for accepting theories
are highly influenced by such application-guided values.

Given the increasing importance of AI systems in society and their potential
benefits and harms, it is perhaps more instructive to understand
(the lack of) value-freeness in Mechanistic Interpretability as we would
in Climate Science or Public Health, rather than in Theoretical Physics.\footnote{
  Some philosophers have further argued that the Value-Free Ideal is not even
  tenable or desirable in Theoretical Physics either.
}
Researchers must share reproducible, quantitative results for the community
to assess but it is unavoidable that the choice of study and what counts as sufficiently
convincing to such and such a conclusion is highly value-laden
\citep{sharkey2025openproblemsmechanisticinterpretability, casper2025pitfalls}.

Expressing an interpretability-flavoured notion of Value-Ladenness,
Dmitry's Koan \citep{dmitrys_koan} states:
\begin{quote}
  \emph{There is no such thing as interpreting a neural network.
  There is only interpreting a neural network at a given scale of precision.}
\end{quote}

We can view Dmitry's Koan as a direct consequence of the Value-Ladenness of explanations
in Mechanistic Interpretability: what counts as a good explanation is highly
dependent on the level of precision that the interpreter desires.
Explanations at a maximal precision would capture lots of noise as well as useful explanatory signal.

For some use cases (e.g., determining the safety of critical AI systems),
higher fidelity explanations may be
required for providing guarantees about model
behaviour and so we would like highly precise explanations.
In other cases, the interpreter may seek sufficient understanding
with which to monitor or steer the model,
which might be possible with a lower fidelity explanation.\footnote{
    \citet{sharkey_sparsify_2024} provide a careful analysis of the trade-offs
    between explanation precision and complexity.
    We also note the close analogy to rate-distortion theory
    in Information Theory.
}
The ideal precision depends on the (human) interpreter's
goals and hence the ideal explanation is inherently value-laden.

Noting that we do not always have an appropriate definition of
what `precision' itself means in our interpreter's context,
we can extend Dmitry's Koan to Dmitry's Koan$^{++}$ \footnote{
  We may also refer to Dmitry's Koan$^{++}$ as Nora's Koan,
  after Interpretability Researcher Nora Belrose who brought this to our attention.
}
stating:

\begin{quote}
  There is no such thing as interpreting a neural network.
  There is only interpreting a neural network at a given scale of precision
  \emph{and a given metric for defining what precision means.}
\end{quote}

Dmitry's Koan$^{++}$ highlights that the choice of precision metric itself is value-laden.

\subsubsection{Theory-Ladenness of Explanations}
\label{sec:theory_ladenness}

We might hope to
understand ML systems on their own terms,
in their ontology, removing all human ``biases'' from the explanation.
After all, if we have enough data, perhaps the data will speak for itself,
we might think.
This (unfortunate) desire is known as the Theory-Free Ideal \citep{andrews2023theory_free}.

Our explanations always contain underlying (human) theory.
Indeed, so-called ``unsupervised'' learning cannot occur
without either pre-defined inductive biases or external supervision
\citep{andrews2023theory_free, wolpert1997no_free_lunch, goldblum2023no_free_lunch, locatello2019unsupervised_disentangled}.
All observations (and interpretations) are theory-laden
\citep{Kuhn1962-strcture_of_sci_rev, Duhem1954-aim_structure_physical_theory, popper_scientific_discovery}.
Interpreter theory seeps into the explanation
in all stages of an interpretability workflow:
from problem formulation and model design to
model selection and semantic interpretation.

\begin{importantbox}[Example: Theory-Ladenness of Sparse Autoencoder Explanations]

    Sparse Autoencoders (SAEs) are a method for unsupervised interpretability,
    aiming to extract concepts (or ``features'') from
    the activation space of neural networks.
    Concept representations are generally
    entangled and difficult to access in the neural activation space.
    We may hope for SAEs to disentangle these representations
    into a linear combination of monosemantic concepts by mapping the neural activations to a feature basis.
    Empirically, it has been shown that disentangled concept representations
    are more amenable to human interpretation
    \citep{anthropic_sae_towards_monosemanticity_bricken}.
    \\

    We might believe that we are doing completely unsupervised
    learning with no human theory
    when producing SAE derived explanations of neural activations.
    However, note that in using the SAE, we are
    committing to the theory that features are sparsely activated
    and linearly represented \citep{anthropic_sae_towards_monosemanticity_bricken}.
    Similarly, in choosing a particular
    SAE architecture like TopK \citep{openai_saes_2024} or Jump-ReLU SAEs \citep{jumprelu_sae_deepmind},
    we are committing to the Monotonic Importance Heuristic \citep{ayonrinde2024feature_choice_saes},
    the conjecture that feature activations are not typically both small and important simultaneously.
    \citet{ekdeep2025projectingsaes} further describe the theoretical assumptions
    that come with different choices of SAE architecture.
    Following \citet{locatello2019unsupervised_disentangled}, we note that
    unsupervised disentanglement learning in the general case is not possible;
    we must first hold some theoretical commitment to the structure of the data.
    We choose theoretical commitments because we have reason
    to believe that they are good inductive priors
    for the data distribution,
    or because we believe that the structures will be more easily
    human-understandable.

\end{importantbox}

Interpretability is not and cannot be a purely engineering
affair, devoid of theory.
We require theory for two reasons:
Firstly, as we have seen with SAEs and disentanglement learning,
holding the wrong theoretical commitments\footnote{
    which is very easy if
    researchers believe that they are holding no theoretical commitments at all!
}
leads to the intractability of unsupervised learning.\footnote{
    Since interpretability aims to understand neural
    systems which we do not yet understand, it is inherently an
    unsupervised learning problem: we don't know
    what we don't know about the system.
}
And secondly, the human (interpreter's)
priors are a key part of the theory, as indeed it's humans that
we would like to make interpretations for!
In this sense, Interpretability is a fundamentally socio-technical
problem which may be best addressed by a combination of understanding
humans, machines and the interactions between the two.
Hence we see a key role for Human-Computer Interaction (HCI) and
the Social Sciences in MI.
We suggest that increased focus on the criteria which make explanations accessible
to humans \citep{schut2023chess_interp,ayonrinde2024_mdl_saes},
especially to diverse humans \citep{diversity_in_hci}, is likely to prove fruitful for
future interpretability work.\footnote{
Theory-ladenness of explanations is also relevant in the context of
the \emph{Construct Validity problem} \citep{cronbach1955construct_validity}
- the problem of whether the explanation is measuring
what it purports to measure.
(As \citet[p.58]{Heisenberg} put it:
``what we observe is not nature in itself but nature exposed to our method of questioning.'')
It is very possible for researchers to agree on the data reported and
the statistical validity of hypothesis tests and yet disagree on the
interpretation because their underlying theories are different
(see also \Cref{sec:types_of_theory} for the use of model and domain theory in interpretability).
}

\subsection{Limits of Model-level vs System-level Explanations}
\label{sec:model_level_limits}

As detailed in \Cref{sec:mech_interp_def},
explanations in Mechanistic Interpretability are inherently Model-level explanations.
However, when interacting with AI,
we are typically interacting with AI \emph{systems}, not \emph{models}.
Though models may think, only systems behave:
it is systems that perform actions in the world.
MI explanations, then, have limited explanatory power
when the system-model relationship is complex or not well-understood.
In systems with meta-decoding processes such as
those with inference-time compute loops, ensembling methods,
or similar, then we might expect
model-level explanations to be insufficient for understanding system-level behaviour.

Systems which can well be described as having an Extended Mind, in the
sense of \citet{Clark1998_extended_mind} or as embodied/embedded agents
\citep{Garrabrant2018embedded_agents, embodied_cognitiion_sep}
may also be difficult to understand with model-level explanations.
It may be difficult to pick out some feature if the
feature as it appears in the model is simply a pointer to some cognitive
process distributed elsewhere in the system.
A particularly notable case of such systems is multi-agent AI systems,
which may have emergent properties not well explained by the analysis of
individual agents.
For example, consider a flock of birds
or a well-functioning marketplace which may not be easily
understood by the analysis of individual agents within the system
\citep{hyland2024freeenergy}.

\subsection{Limits of Low Abstraction Explanations}
\label{sec:low_abstraction_explanations}

One other possible, though surmountable, limit to
Mechanistic Interpretability explanations is that low-level
explanations may be difficult, or even impossible, to turn into
explanations at higher levels of abstraction.\footnote{
    Note that by levels of abstraction here we do not mean the
    semantic abstraction level of a given feature, where some features
    might represent more higher-level concepts than others (for example those
    at later layers of a model). We instead mean that features themselves
    are low-level units compared to higher-level abstractions like
    components or circuits which they may compose.
}
For example, in the natural sciences, it is not generally known
whether the laws of thermodynamics can be derived from lower-level
particle physics laws.
If MI provides low-level explanations akin to quantum mechanics,
research questions that more closely resemble chemistry, biology or, social science questions
may not be obviously derivable in a reductionist way from MI explanations.

\section{Discussion}
\label{sec:discussion}

Given a data distribution $\mathcal{D}$, suppose that we would like to
explain the behaviour of a neural network $M$ over a subset
of the data distribution $D \subset \mathcal{D}$.
Then the goal of Mechanistic Interpretability is
to provide a Model-level, Ontic, Causal-Mechanistic, Falsifiable explanation $E$
of the model's behaviour on $D$ which is explanatorily faithful to $M$.
Where behavioural faithfulness requires that the end predictions of the models agree,
\emph{explanatory faithfulness} is a stronger condition that requires
the causal structure
of the explanation $E$ to be faithful to the causal structure of $M$
at each stage of the causal chain.

In this work, we have argued that approaching the Problem of Interpretability
through the Explanatory View of Mechanistic Interpretability is likely to
be fruitful because neural networks naturally admit explanations
of their behaviour through their internal structures.
Hence, our explanatory methods can and should look to
uncover causal structure in the model $M$
rather than merely producing confabulatory descriptions
of model behaviour.
The Explanatory View of Neural Networks provides a justification
for using explanatory faithfulness as the goal of explanations
in Mechanistic Interpretability.

However, there are significant limitations to this approach.
In particular, it is infeasible to have a general algorithm for finding
explanations $E$ for all models $M$ and all data distributions $D$
which are optimal for all purposes.
Solutions to the interpretability problem are Theory-Laden:
they require some theoretical priors about neural networks
and/or the data distribution to find good explanations.
Similarly, the problem of finding good explanations is Value-Laden:
what makes for a good explanation depends on our goals as interpreters.

A core problem to address in future work is how to appropriately
characterise what makes a \emph{good explanation} in the context of Mechanistic
Interpretability.
Here, we have argued for necessary criteria for an explanation
to be validly considered `Mechanistic' (namely that it is Model-level, Ontic, Causal-Mechanistic
and Falsifiable).
We would further like to understand how some
explanations are better than others in terms of their usefulness
and likelihood to point towards truth.

\hypertarget{explanatory_optimism}{
\section{Coda: Explanatory Optimism}\label{sec:explanatory_optimism}}

\begin{quote}
  \itshape ``Have you persuaded yourself that there are knowledges and truths beyond your grasp,
  things that you simply cannot learn?
  ...
  If you have allowed this to happen,
  you have arbitrarily imposed limits on your intellectual freedom,
  and you have smothered the fires from which all other freedoms arise.''
  \par\raggedleft\normalfont --- Scott Buchanan, 1958
  \end{quote}

We conclude by introducing a conjecture at the heart of (Mechanistic) Interpretability,
that we call \textbf{The Principle of Explanatory Optimism}.
We would like to explicate this conjecture,
argue for its importance for MI research, and
raise a Call to Action for further research into
clarifying both the statement and its veracity.
Here we provide no arguments for
the truth of Explanatory Optimism (EO) as a conjecture;
we leave further arguments, proofs, or refutations to future work.

\subsection{Alien Concepts}

Suppose that an AI \textcolor{blue}{M} (Machine, in blue) and the
interpreter \textcolor{red}{H} (Human, in red)
each have some set of concepts which are understandable to them,
$\textcolor{blue}{C_M}$ and $\textcolor{red}{C_H}$ respectively
\citep{schut2023chess_interp, hewitt2025ai_vocabulary}.
If $\textcolor{blue}{C_M} \subset \textcolor{red}{C_H}$, then intuitively
all concepts that the machine uses for its computations
can be immediately understood by the human interpreter.
However, if $\textcolor{blue}{C_M} \setminus \textcolor{red}{C_H}$ is large,
then there are Machine-concepts that are not natively Human-understandable.

\begin{figure}[h]
  \centering
  \begin{minipage}{0.5\linewidth}
      \centering
      \includegraphics[width=\linewidth]{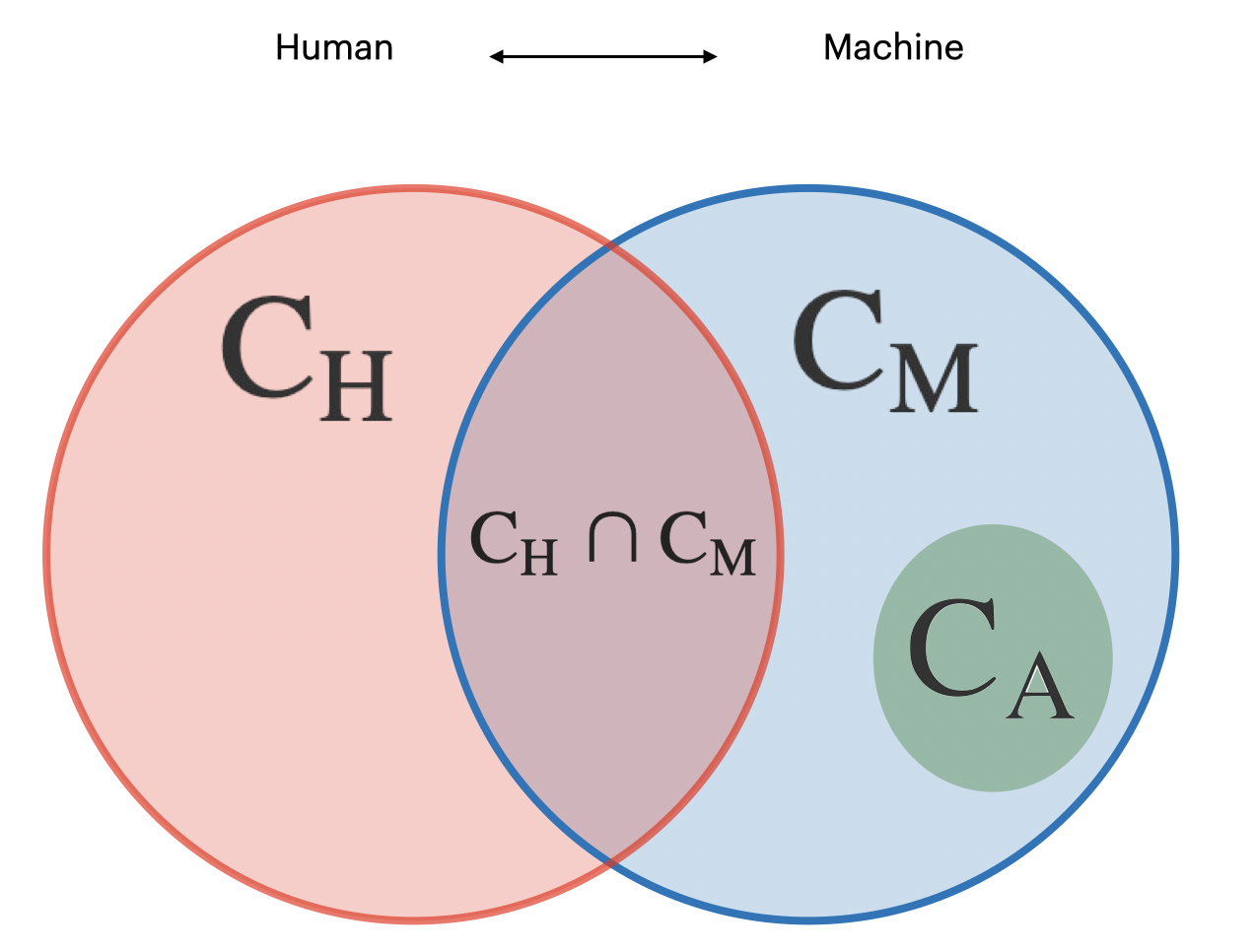}
  \end{minipage}
\caption{
  A Venn diagram showing the relationship between the
  concept spaces of the machine \textcolor{blue}{M} and human interpreter \textcolor{red}{H}.
  The machine and human have some shared concepts which they can use to communicate
  ($\textcolor{blue}{C_M} \cap \textcolor{red}{C_H}$)
  but there are many concepts that the machine uses that the human does not understand
  ($\textcolor{blue}{C_M} \setminus \textcolor{red}{C_H}$).
  The set of \emph{Alien Concepts}
  $\textcolor{green!60!black}{C_A} \subset (\textcolor{blue}{C_M} \setminus \textcolor{red}{C_H})$,
  is a subset of the Machine-concepts.
  Alien Concepts are causally relevant for the model's computation but are fundamentally incomprehensible to humans.
  If this set is large or important, then Interpretability may be highly limited.
}
\label{fig:alien_concepts}
\end{figure}

Some Machine-concepts that aren't intuitively understandable by humans
may be human-understandable with some human effort,
effective translation, or good explanations.
However, a core problem remains if there are concepts that are understandable to
the model but which are fundamentally alien and incomprehensible to humans.
We call such concepts \textbf{Alien Concepts} and label these $\textcolor{green!60!black}{C_A}$.
Alien concepts are Machine-concepts in $\textcolor{blue}{C_M} \setminus \textcolor{red}{C_H}$
that are effectively untranslatable into Human-concept terms (see \Cref{fig:alien_concepts}).

To the extent that Alien Concepts are present in the model,
and are important for the model's computation,
the project of Interpretability may be fundamentally flawed.
The conjecture at the heart of (Mechanistic) Interpretability then is that artificial neural
network-based general intelligences have few, or no, alien concepts
that are both vital to model behaviour and not human-understandable.
A version of this conjecture appears to
be a prerequisite for interpretability research:
if most of the model's concepts aren't understandable to MI
researchers, then it is not clear how MI research can proceed.\footnote{
  Conversely, if model concepts \emph{are} possible to be made understandable
  by MI researchers,
  then MI research is a tenable research direction
  (though it may still be difficult).
}

\subsection{The Principle of Explanatory Optimism}

As phrased above, the ``no alien concepts'' conjecture is about neural networks.
It may be more instructive, however, to rephrase this conjecture to put human intelligence at the center
and understand it as a claim about a class of general intelligence
(such as neural-network based AI systems).
Rephrasing then:
\emph{Everything that is important for the behaviour of an intelligence with implicit explanatory
knowledge within some explanatory complexity class is human-understandable.}
This statement is a strong form
of the conjecture that we call
\textbf{The Principle of Explanatory Optimism}.\footnote{
    This idea is closely analogous to
    \citet{deutsch2011beginning_of_infinity}'s theory of Optimism.
}
Weaker forms of Explanatory Optimism might claim that ``most'' rather than all neural
network behaviour is human-understandable (in terms of variance explained for example).

Humans have certain cognitive limitations
compared to future generally intelligent systems:
we have limited memory and processing power, we are somewhat limited in bandwidth and speed,
and we may lack attention.
Strong Explanatory Optimism suggests that, given sufficient time,
we could understand these artificial intelligences,
\emph{if} augmented with good, concise explanations,
memory devices, cognitive tools and such like.
The Explanatory Optimism conjecture implies that
explanations for understanding Machine-Concepts exist.

\begin{optimismbox}[The Principle of Explanatory Optimism]
  \textbf{The Strong Principle of Explanatory Optimism (SEO)}:
  \begin{quote}
    \itshape
    Everything important for the behaviour of an intelligence with implicit explanatory
    knowledge within some explanatory complexity class is human-understandable.
  \end{quote}

  \textbf{The Weak Principle of Explanatory Optimism (WEO)}:
  \begin{quote}
    \itshape
    Most important behaviour of an intelligence with implicit explanatory
    knowledge within some explanatory complexity class is human-understandable.
  \end{quote}

\end{optimismbox}

Explanatory Optimism (EO) can also be understood, as in \citet{deutsch2011beginning_of_infinity}, as a view of
explanatory universality, defined analogously to computational universality as
given in the Church-Turing thesis \citep{turing1936computable,church1936unsolvable}.
In the same sense that any Turing machine can simulate any other Turing machine,
we would like a theory that maintains that some intelligences are
explanatorily universal in the sense that they can understand and explain any other intelligence
of an equivalent explanatory complexity class.
We leave the question of what such an explanatory
complexity class might look like and how to prove explanatory
class equivalence and universality to future work.

The truth value of Weak Explanatory Optimism is a load-bearing question for the field of interpretability:
if models aren't human-understandable, then the field will face unassailable
roadblocks in its mission to explain neural networks to humans.
Hence, an appropriate disproof of, or otherwise sufficiently convincing arguments against,
(W)EO should motivate researchers working on Mechanistic Interpretability
to consider reorienting their research focus.\footnote{
    MI researchers with a
    downstream goal of AI Safety, AI Ethics, or AI Cognitive Science
    may be interested in whether EO holds.
    Absent EO, MI may not be an effective way to reach their goals.
}

\begin{optimismbox}[Call to Action for Explanatory Optimism]
  The Call To Action for future work is twofold:
  \begin{itemize}
    \item Firstly, to formalise the above conjectures (SEO and WEO).
        We would like to develop core definitions
        for the explanatory complexity classes and the appropriate notion of explanatory universality.
        This formalisation would likely involve
        understanding the explanatory complexity classes of different intelligences
        as well as operationalising the quantitative notion of understanding
        ``most'' of a model's behaviour.

    \item Secondly, to prove the formalised conjectures.
        We would like to assess the
        truth value of the Principle of Explanatory Optimism.
        We believe that such results would be of great interest,
        both to interpretability researchers and
        scientists who expect to use non-transparent computational models.
  \end{itemize}
\end{optimismbox}

Complexity theorists, theoretical computer scientists,
analytic philosophers, and computational mechanics theorists
may be able to use the tools from their fields to make progress on this
conjecture.\footnote{
    Explanatory Optimism is another area where we might expect
    fruitful collaboration between Philosophy and Computational Complexity
    (and Machine Learning) as in \citet{Aaronson2013-philosophers_computational_complexity}.
}

\subsection{The Importance of Explanatory Optimism}

As ML models begin to do more cognitive work in the world,
the frontiers of knowledge in
mathematics, the (natural, social, and
computational) sciences, and the humanities may be
known to machines before humans.\footnote{
    Arguably AI automation of science may be the final
    stage of the long-continuing
    Crisis of the European Sciences depicted by
    \citet{husserl1970crisis_of_european_sciences}.
    For Husserl, the fact that the sciences have become so
    disconnected from the phenomenological world of everyday experience
    results in scientific disciplines becoming increasingly specialised
    such that no human has a unified understanding of science as a whole.
    With sufficiently intelligent AI, we can imagine a world where
    no human has an understanding of the furthest advances in
    \emph{even a single scientific discipline}.
}
The default state of the world when living alongside such cognitively
advanced machines is that humans are \emph{epistemically disempowered}
and subjected to living in a world built by knowledge that no human understands.
However, Explanatory Optimism offers an alternative future for humanity.
The upshot of Explanatory Optimism is that as machines learn more about the world,
through interpretability, humans can learn more about the world too.
Any Machine Knowledge can become Human Knowledge.

Hence, if Explanatory Optimism is true, Interpretability
may be one of the most important projects in the history of modern science.
Explanatory Optimism implies that \emph{all explanatory knowledge is accessible to people}
through interpretability and human-computer interaction:
we are sitting at but the beginning of an explosion in human understanding.

\newpage

\hypertarget{acknowledgments}{
\subsubsection*{Acknowledgments}\label{acknowledgments}}
Thanks to Nora Belrose, Matthew Farr, Sean Trott, Elsie Jang, Evžen Wybitul, Andy Artiti,
Owen Parsons, Kristaps Kallaste and Egg Syntax for comments on early drafts.
We appreciate Daniel Filan and Joseph Miller's helpful feedback.
Thanks to Mel Andrews, Alexander Gietelink Oldenziel, Jacob Pfau,
Michael Pearce, Catherine Fist, Lee Sharkey,
Jason Gross, Joseph Bloom, Nick Shea,
Barnaby Crook, Eleni Angelou, Dashiell Stander, Geoffrey Irving and
attendees of the ICML2024  MechInterp Social for useful conversations.
We're grateful to Kwamina Orleans-Pobee, Will Kirby and Aliya Ahmad for additional support.
This project was supported in part by a Foresight Institute AI Safety Grant.

\newpage

\bibliographystyle{colm2025_conference}
\bibliography{references}

\newpage
\appendix

\hypertarget{appendix_examples}{
\section{Examples of Explanation Types}\label{sec:examples}}

In this section, we provide some intuitive examples and non-examples of Explanations
which satisfy the criteria that we outline in \Cref{sec:mech_interp_def}.

\subsection{Ontic Explanations}

\emph{Question}: Why did the pen fall off the desk?

\paragraph{Causal-Mechanistic But Not Ontic Explanation.}
\begin{quote}The pen fell off the desk because the aether pushed the bottle and then the bottle pushed the pen off the desk.
\end{quote}
This explanation is Causal-Mechanistic in the sense that one thing happens after another and causes the next.
However, if we do not believe that the aether is a real entity then this explanation cannot be considered an Ontic Explanation.

---

\emph{Question}: Why is the cube heavy?

\paragraph{Ontic But Not Causal-Mechanistic Explanation.}
\begin{quote}The cube is heavy because it is made up of tungsten atoms. \end{quote}
This explanation is Ontic as the entities involved in the explanation are
real entities.
However, it is not Causal-Mechanistic as there is no step-by-step explanation without gaps.

\subsection{Statistically-Relevant Explanations}

Consider the explanation:
\begin{quote}
    Ice cream sales are higher on days when there are more shark attacks.
    If there's a shark attack reported,
    we can predict with 85\% confidence that ice cream sales will be above average that day.
\end{quote}
This explanation is purely in terms of statistical correlation rather than causation.
There is no explication of any underlying causal mechanism,
which might involve both phenomena being causally downstream of hot
weather and/or more beach visitors.
We could perform interventions to test this hypothesis.

\subsection{Telic Explanations}

Consider the following explanation:
\begin{quote}
    The heart exists to pump blood throughout the body and maintain circulation.
\end{quote}
This explanation describes the purpose and function of the heart,
rather than describing the physical processes by which the observed phenomena
of blood pumping comes about (which might include a mechanistic description of
chambers, valves and muscles).
In this way, this explanation does not provide a
continuous causal chain from cause to effect, without any unexplained gaps,
as a Causal-Mechanistic explanation should.

\subsection{Nomological Explanations}

\emph{Question}: Why does a metal rod expand when heated?

\paragraph{Nomological but not Causal-Mechanistic Explanation.}
\begin{quote}
    The rod expands because it follows the natural law that all metals expand when heated,
    as described by the coefficient of thermal expansion.
\end{quote}
This explanation references a general law of nature without getting into the underlying mechanism.

\paragraph{Causal-Mechanistic Explanation.}
\begin{quote}
    The rod expands because
    its metal atoms vibrate more vigorously when heated, which increases their average spacing.
    This increased spacing leads to an overall increase in the rod's length.
\end{quote}
This details the physical mechanism causing the expansion.

\hypertarget{appendix_nma}{
\section{The No Miracles Argument for the Explanatory View}\label{sec:nma}}

In \Cref{sec:mech_interp_def}, one of our conditions for claiming that scientific explanation
of a neural network was a valid Mechanistic Interpretability explanation was that the
entities in the explanation are (to the best of our knowledge) real (Ontic) entities.
We might wonder how it is possible to make such a claim, given that it is not
necessarily clear that there's an ontic test for features within a neural network.

Similarly, in discussing representations, following \citet{harding2023representations_nlp},
we noted that Information, Use and the possibility for Misrepresentation are required
to claim that some activations are \emph{representations}. However, it is not entirely clear
that we are able to draw a correspondence between the world and the activations (i.e., to say that
the neural network is really \emph{representing} a feature of the world or that the feature
is \emph{about} the world).

Our notion of ur-explanations, defined in \Cref{sec:what_we're_explaining} as
idealised explanations of model behaviour on an input distribution,
given in terms of its learned internal structures, seems to also rely on representations
being appropriately well-defined as above.

We here provide an argument that
mitigates the above concerns.
This argument will go via the traditional No Miracles Argument (NMA) for Scientific Realism
\citep{Putnam1979-putnam_nma, Lipton1994_truth_ibe, Psillos1999_scientific_realism, rowbottom2024_nma_for_ai}.
We will first provide a brief overview of the NMA in Science,
then adapt the NMA for Machine Learning models in general,
and finally we will adapt the NMA for the explanations
of Mechanistic Interpretability.

\subsection{The No Miracles Argument}

\begin{quote}
    \itshape ``(Novel) empirical successes in science enabled
    by scientific theories are non-miraculous because such theories
    are typically or probably approximately true.''
    \par\raggedleft\normalfont --- \citet{Putnam1979-putnam_nma}
\end{quote}

The No Miracles Argument (NMA) is widely considered to be the strongest argument for
Scientific Realism and contends that
``[Scientific Realism] is the only philosophy that
doesn’t make the success of science a miracle''
\citep{Putnam1979-putnam_nma}. The argument proceeds as follows \citep{sep-scientific-realism}:
\begin{enumerate}
    \item Scientific theories are (extraordinarily) successful in the sense that they make accurate
        novel empirical predictions about phenomena of interest.
    \item If our scientific theories are very far from the truth, then it would be miraculous that
        they are so successful.
    \item Given the choice between a straightforward reason for the success of scientific theories
        and a seemingly miraculous sense in which all our theories are just coincidentally producing
        accurate novel predictions, one should clearly prefer the former.
    \item Therefore, two conclusions follow:
        \begin{enumerate}
            \item Firstly, that our best scientific theories are approximately true (or approximately
                correctly describe mind-independent laws)
            \item Secondly, that the entities posited by such scientific theories are real (or approximately
                characterize mind-independent entities).
        \end{enumerate}
\end{enumerate}

Note that the NMA has two distinct conclusions.
The first conclusion is the \emph{epistemic thesis} which states that
our best scientific theories are approximately true.
for example, that the fact that the effective predictions of Relativity enable
GPS-based navigation systems provides good reason to believe that
Relativity is approximately true.
The second conclusion is the \emph{semantic thesis} which states that
scientific entities \emph{refer}.
For example, that when physicists talk about electrons that they have not
seen with their naked eye that they are referring to real entities.

\citet{Dawid2016_nma_without_base_rate_fallacy} provide a formalisation of the No
Miracles Argument in terms of Bayesian probability which we refer readers
to for a complete mathematical formalism of the NMA.\footnote{
    Note that \citet{Dawid2016_nma_without_base_rate_fallacy}'s formalisation
    avoids an earlier mistaken formalisation which contained a base rate fallacy error.
    See also \citet{howson2000hume} for further discussion.
}

\subsection{The No Miracles Argument for Neural Representations}

We would like to provide an argument that neural activations in
well-trained, generalising neural networks, are representations
in the sense that they correspond with entities in the world.

From the Explanatory View of Neural Networks (see \Cref{sec:induction_to_explanation}),
we see that the NMA can be applied to the case of neural networks.
In the context of science, neural networks often play the role of theory:
they are concise representations produced by processing data,
which, at inference time, provide predictions for new data.
In this sense neural networks play a theory-like epistemic role.
We may hence insert the neural networks into the role of theory in the NMA as follows.

First, we note that neural networks in general are extraordinarily successful at making
accurate predictions.
This success occurs both for individual narrow tasks like
image classification \citep{lecun1998gradient} and playing Go \citep{schrittwieser2020alpha_zero}
as well as for general tasks like language modelling \citep{brown2020gpt3} and
writing code \citep{li2022_alpha_code}.

We may now follow the NMA's argument:
if the representations and computations of neural networks
don't at least approximately correspond to entities and laws in the training data
(and thus in the natural world) then their quite incredible success would be
miraculous. The fact that neural networks generalise so well, that is they make
accurate novel predictions, suggests that we have good reason to believe in
the non-miraculous conclusion - the structure in neural networks produces
representations that correspond to the entities in the world
(the semantic thesis of the NMA).
And we ought to think of the explanatory theories implicit within the neural network
as providing a useful guide to approximate theories about the scientific structure
of the task (the epistemic thesis of the NMA).\footnote{
    Also note the implications here for interpretability
    as a tool for learning about science through scientific AI models.
}

\subsection{Explanatory Faithfulness Through the No Miracles Argument}
\label{sec:nma_explanatory_faithfulness}

Our application of the NMA to neural networks suggests that ur-explanations
provided by neural networks are scientifically interesting
explanatory theories which are increasingly likely to capture
the true structure of the world (asymptotically closely) as models
become better predictors.\footnote{
    This argument could provide an explanation for the Platonic Representation
    Hypothesis of \citet{pmlr_v235_platonic_representations},
    which suggests that the representations of neural networks
    tend to become more similar as the networks become larger and
    more effective predictors. The NMA argument suggests that they
    are converging to the true common structure of the data
    generating process.
}

However we also note that in practising Mechanistic Interpretability,
researchers come up with their own explanations of model behaviour.
These researchers would like to uncover the ur-explanations but
a priori they have no way of knowing that their explanations
coincide with the idealised ur-explanations that they are targeting.
However, a repeated application of the NMA can provide such evidence.
We may call the following argument \textbf{NMA}$^\mathbf{2}$.

\emph{Firstly}, the No Miracles Argument as given above gives us reason
to believe that there are representations and ur-explanations
to be found always-already within the trained neural network.
This is simply a straightforward application of the NMA.

\emph{Secondly}, suppose that we have a mechanistic explanation, $E$ of a neural network.
It is theoretically reasonable to ask whether the explanation, $E$, is explanatorily
faithful to the ur-explanation, $U$.
However, it is not immediately clear how to
practically evaluate the claim of whether $E$ is explanatorily faithful to $U$.
We note however that we may again apply the NMA to the explanation $E$.
That is to say, if (1) the explanation $E$ highly successfully predicts the
relevant information in the neural network activations at each layer of the network,
and (2) the activations are representations in the sense of \citet{harding2023representations_nlp}
under causal interventions, then we can conclude (C) we have reason to believe that $E$
in fact does approximately correspond to the ur-explanation $U$.\footnote{
    An example of an explanation that we might consider explanatorily faithful in the above sense
    is the explanation of Modular Arithmetic in \citet{nanda2023grokking}.
    Some other MI explanations are less successful layer by layer and incur non-trivial
    change to the performance of the model if patched as a replacement for some of the
    model's computation.
}
It would be highly coincidental (miraculous even) for the negation to be
routinely the case. (Here again we use the epistemic thesis of the NMA.)

As a practical upshot, this argument provides some validity to the methods
of \citet{shi2024hypothesis_testing_circuits}, who propose
hypothesis testing as a method for evaluating the faithfulness
of mechanistic explanations.
If the circuit explanation $E$ passes the hypothesis test,
at an appropriate significance level,
then we have reason to believe that $E$ similarly corresponds
to the ur-explanation $U$; that is $E$ is not merely behaviourally
faithful but further is explanatorily faithful.\footnote{
    In practise we might consider the significance level
    of the tests that are run in practise to provide
    much too weak evidence to conclude that the explanation
    is explanatorily faithful.
}

In summary, through the \textbf{NMA}$^\mathbf{2}$ argument, we believe that
the following two conclusions follow:
\begin{enumerate}
    \item We have reason to believe that ML models that are
        extraordinarily successful at making accurate novel predictions
        contain explanatory knowledge.
        That is, their ur-explanations
        are well-defined and we ought to be realists about features,
        as variables of computation.
        Further, ur-explanations are to be unique.

    \item We have reason to believe that MI explanations that provide
        successful algorithmic predictions of how the neural activations
        are structured at each sequential layer of the model,
        are likely to be approximately explanatorily faithful
        to the model's ur-explanation
        and contribute to our understanding of the model's behaviour.

\end{enumerate}

\hypertarget{appendix_mi_value}{
\section{The Value of Demarcating Many Valuable Forms of Interpretability}\label{sec:appendix_mi_value}}
In \Cref{sec:mech_interp_def}, we provided a technical definition
of Mechanistic Interpretability (MI) as producing explanations
of neural networks that are \emph{Model-level},
\emph{Ontic}, \emph{Causal-Mechanistic} and \emph{Falsifiable}.
This definition provides a clear demarcation of interpretability research
which ought to be called Mechanistic Interpretability and
research which falls outside of Mechanistic Interpretability.

In making this distinction, we attach no value judgements to the term ``Mechanistic''.
We are not wanting to say that Mechanistic Interpretability
is inherently good or uniquely valuable
and that non-Mechanistic is necessarily bad or worthless, or vice versa.
Demarcating Mechanistic Interpretability is simply a useful way
to understand what is meant by the term
in a way that is useful for us to make claims about the efficacy of
certain methods to solve the interpretability problem.
It seems highly plausible that
Chain-of-Thought Interpretability (non-Model-level),
Concept-Based Interpretability (non-Causal-Mechanistic), and
Propositional Interpretability (non-Causal-Mechanistic)
could be practically useful
for downstream tasks or as inputs to Mechanistic Interpretability.

Though many flowers should bloom, it is prudent for us
to refrain from calling daffodils tulips.

\hypertarget{three_varieties}{
\section{The Three Varieties of Interpretability}\label{sec:three_varieties}}
Under the classical (behavioural) view of Machine Learning, there
are two ways to have an explanation of a model's behaviour:
\emph{Interpretability By Design} and \emph{Post-hoc Interpretability} \citep{lipton2018mythos_classical_interp}.
Linear models and decision trees are examples of models that are considered interpretable\footnote{
    The interpretability of such models is somewhat subjective and
    not solely dependent on the high-level architecture.
    Very large high dimensional models, even if they are simple,
    may be difficult to interpret.
    See \citet{izza2020explaining_decision_trees} for discussion of decision trees
    and \citet{lipton2018mythos_classical_interp} for a discussion of linear models.
    In particular, when the features of a linear model are highly correlated the
    interpretability of the model may be reduced.
}
by design:
it is typically considered relatively easy to understand how these models make their
predictions by inspection.\footnote{
    Interpretable by Design models may
    suffer a performance penalty for their interpretability.
    We might think of this
    performance degradation as an \emph{Interpretability Tax},
    a cost paid for the ability to understand the model's behaviour.
    See also the Alignment Tax in AI Safety
    \citep{lin-etal-2024_alignment_tax}.
}
Post-hoc interpretability methods, on the other hand, are methods which take a trained model and
attempt to confer useful information by creating explanations which statistically
seem to correlate well with the model's behaviour.\footnote{
  One interesting form of post-hoc interpretability is
  models that attempt to self-explain.
  For example, a language model might output a Chain of Thought,
  either after the fact or as part of the generation process.
  However, in general, models do not reliably and faithfully describe
  the process underlying their outputs
  \citep{turpin2023unfaithful_cot, lanham2023faithfulness, chua2025r1_faithful,chenreasoning_model_faithfulness}.
  We may hence think of these Chain of Thought ``explanations''
  as epiphenomenal rather than faithful explanations.
}

\emph{Mechanistic Interpretability} offers a third variety of explanation distinct from both
Interpretability By Design and Post-hoc Interpretability.
With the Explanatory View of neural networks,
we can make sense of a new type of interpretability functioning as a pursuit to \emph{uncover}
the ur-explanations that are always-already present within the trained network.

The Explanatory View of Neural Networks takes seriously
the idea that there is structure in the
model to be interpreted.
We can contrast this with post-hoc interpretability
where we may come up with just-so stories that happen to match the network's behaviour
on a sub-distribution,
even if they are confabulatory and not explanatorily faithful.
The Explanatory View is a cognitive
realist view of neural networks that suggests that there
exists a target to the interpretability program:
we are not merely looking for explanations which appear to correlate with model behaviour,
models admit explanations that we would like to extract.

\hypertarget{appendix_types_of_theory}{
\section{The Theory Required For Interpretability}\label{sec:types_of_theory}}
\Cref{sec:theory_ladenness} provides a discussion
of theory-ladenness in interpretability methods.

There are two plausible types of theory that we might seek
to understand a domain-specific neural network,
\emph{Domain Theory} and \emph{Model Theory}.
For example, if we seek to interpret a protein model then the
Domain Theory concerns the biology of proteins.
Model Theory, however, always concerns neural networks.

All interpretability requires Model Theory: it would be useful
to understand the structure of neural networks, how they generalise,
how they represent the structure of the world, etc.
The activation space is a function of the preceding weights of the model
and the input data distribution, however understanding activation spaces
isn't inherently a function of the Domain Theory.
It is not the case that to interpret a protein model, we should need
lots of Domain Theory;
we need only minimal knowledge about proteins ahead of explanation.
On the contrary, we should hope that using interpretability we can
learn more about the Domain Theory, from the network.
That is, a scientific benefit of interpretability research on
scientific models (e.g. protein models) is that they can help
us to learn more about Domain theory (e.g. protein theory) than we knew before
through understanding the model's knowledge of the subject matter.

\hypertarget{straightforward_explanation}{
\section{The Straightforward Implementation-Level Explanation}\label{sec:straightforward_explanation}}
Following \citet{marrs_levels}, in \Cref{sec:computations_over_representations},
we discussed the distinction between the Computational, Algorithmic
and Implementation levels of analysis.

All neural networks admit at least one trivial explanation at the Implementation leval
which we denote the \emph{straightforward explanation}.
We define the straightforward explanation as follows.
Given a neural network $f:X\to Y$ and $x\in X$ such that $f(x) = y\in Y$.
The straightforward explanation is given by a trace of $f(x)$,
the computation of the network on the input $x$.
In fact, this explanation is a formal proof of the equality $f(x)=y$.

While straightforward explanations in this sense are always available,
these are typically not \emph{good} explanations
as they are not very concise or illuminating.
We would like explanations of neural networks that are in terms of the
features (or concepts) that the network learned during training
and explanations which are compact and useful.

\hypertarget{appendix_oracles}{
\section{The Trouble with Oracles and Zero-Knowledge Proofs}\label{sec:appendix_oracles}}
In \Cref{sec:induction_to_explanation}, we described the classical
view of Machine Learning as analogous to seeing the
ML model as a black-box providing \emph{explanationless predictions}
as an oracle would provide prophecies.
We would like to trust models due to the content of their explanations
(that is their reasons for acting and the interventions which
would have changed their prediction),
which is not possible if we have only explanationless predictions.

We may especially desire content-relevant reasons to trust models
in high-stakes deployments of ML-based systems relevant to AI Ethics and
AI Safety \citep{Amodei_2025_urgency_interpretability}.
Similarly, if we would like to do science with ML models,
i.e., generating explanatory knowledge of natural phenomena,
then this appears untenable without understanding the relevant
contentful reasons for the model's predictions
\citep{Raz2024-importance_of_understanding}.

Suppose, for example, that an AI tells a user that if they invest in a certain
stock they will make a lot of money.
To act on this advice, the user may like to know the reasons that the
stock is likely to be a good purchase in terms of the business fundamentals
(an explanation of the prediction) rather than extrinsic reasons (e.g.
the AI has made successful predictions before).
This is especially important when the AI might have different motivations
for the suggestion or may be scheming to mislead the user towards
its own aims.\footnote{
    That is, we don't just face the generic problem of induction, but further, the system
    might be genuinely adversarial and anti-inductive!
}

A core problem with behaviourist explanations in general is that the input space
is too large for the explanation to pick out every single case and hence either (1) the
explanation will be incomplete, leaving out large sections of the input data domain,
or (2) will coarse-grain over the input data domain.
In both cases, these explanations are likely to be insufficient for out of distribution behaviour
and would not be recommended for high-stakes explanations that we would like to be
confident in.

Analogously, zero-knowledge proofs can be thought of as another
form of explanationless prediction - they can convince the listener
of the truth of some statement whilst providing the listener
with no understanding or useful explanatory knowledge.
The reasons to believe the statement of which we have a zero-knowledge proof
do not come from the relevant content of the explanation.
Good explanations are compressions, not cryptography.

\hypertarget{conceptual_engineering}{
\section{Conceptual Engineering \emph{for} and \emph{by} Neural Networks}\label{sec:conceptual_engineering}}
Early Machine Learning models typically required researchers to first
do feature engineering. Before inputting the data into a model,
researchers would transform the input data to
make it more amenable to the model \citep{zheng2018feature_eng}.

\begin{figure}[h]
  \centering
  \begin{minipage}{0.9\linewidth}
      \centering
      \includegraphics[width=\linewidth]{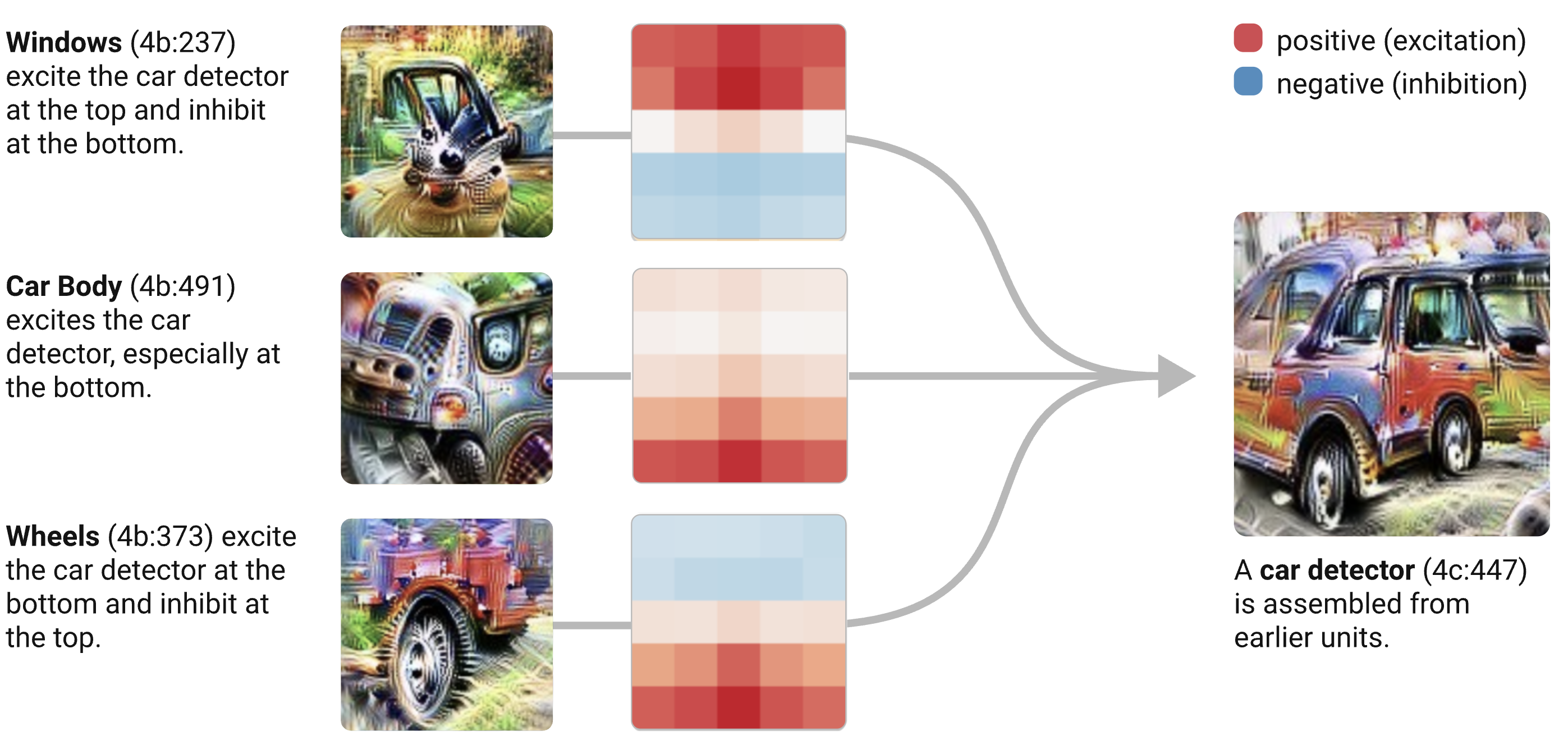}
  \end{minipage}
\caption{
  When studying InceptionV1 \citep{szegedy2015inceptionv1}, \citet{olah2020zoom_in_circuits}
  note that though the network was trained only to classify ImageNet images \citep{russakovsky2015imagenet},
  the network learned intermediate representations that were useful for the image
  classification task.
  For example, in order to detect cars, the network internally learned concepts
  for windows, wheels and car bodies.
  This is an example of \emph{Conceptual Engineering} in neural networks.
  [Image from \citet{olah2020zoom_in_circuits}]
}
\label{fig:conceptual_engineering}
\end{figure}

Modern deep neural networks do not require feature engineering.
Instead, neural networks learn to create representations of the raw input data
which are useful for the task at hand,
without being explicitly told to create such and such a representation
\citep{zeiler2014feature_learning, lecun1998gradient, lecun2015deep_learning}.
For example, \citet{olah2020zoom_in_circuits} find that neural networks trained
to classify images of animals and cars learn to represent the concepts of ``fur'' and ``car windows''
in their intermediate layers (see \Cref{fig:conceptual_engineering}).

Analytic philosophers use the term \emph{Conceptual Engineering} to describe the process
of designing, improving and assessing the concepts that we use in order to better
achieve our aims \citep{Creath1990-carnap_conceptual_engineering, chalmers2020conceptual_engineeering}.
In this sense, generalising neural networks engage in conceptual engineering throughout training:
they learn representations which more closely represent the useful concepts required
for their environment and/or task.
This process of learning representations is Conceptual Engineering
both \emph{for} and \emph{by} neural networks.

In this sense, when we speak about \emph{ur-explanations} of a neural network
as \emph{internal computations} over \emph{learned representations}
(see \Cref{sec:computations_over_representations}),
the representations we are considering have been engineered and learned by
the network at train time in a process of automated Conceptual Engineering.

\newpage

\hypertarget{appendix_universality}{
\section{Explanatory Universality}\label{sec:appendix_universality}}

In \Cref{sec:explanatory_optimism}, we defined Explanatory Optimism as the
conjecture that there exists explanations that explain the behaviour
of an intelligent system (within some explanatory complexity class)
such that a human could understand the system's behaviour.
For a human to understand such an explanation, we should
expect that the explanation can be written in natural
language or some other human-readable format and that the
explanation is relatively short -
say its description length is a sublinear
function of the number of parameters
(and crucially this function should not depend on the
number of input examples which is often very large).\footnote{
    While it may be computationally expensive to find such an explanation
    (that is interpretability may be difficult),
    the claim of Explanatory Optimism is that
    these explanations are available.
    We might expect that arguments for Explanatory
    Optimism thus look more like existence proofs
    rather than a construction of a concise explanation.
    Once someone has found a good explanation of a neural
    network though, by whatever means,
    we should expect that this explanation
    can be understood by a human, given enough time and memory.
}

We here rephrase Explanatory Optimism in terms of
two types of universality that we might look to understand within
neural networks: \textbf{Universality of Learned Concepts}
and \textbf{Universality of Learnable Concepts}.

\subsection{Universality of Learned Concepts}

A core problem
in interpretability is that there is evidence that humans and machines
do not necessarily share the same concepts and representations \citep{hewitt2025ai_vocabulary}.
This discrepancy leads to a problem for interpretability in communicating without
a shared conceptual framework \citep{hewitt2025ai_vocabulary}.

\emph{Universality of Learned Concepts} states that, at sufficient
scale, all neural systems develop universal concepts and mechanisms
(perhaps under the constraint of having a shared environment/training distribution
or having a shared task).
We could either understand this claim by way of networks of sufficiently low prediction error
learning the same set of core concepts or by way of concepts being learned in the same
order,\footnote{
    after some initial period of architecture dependent learning due to the
    initialisation and inductive biases of the network
}
say from simple to complex \citep{belrose2024complexity, rende2024simplicity_bias,hutter2023aixi}.

Universality of Learned Concepts is the sense of universality which
is analysed in interpretability research such as
\citet{olah2020zoom_in_circuits,chughtai2023group_universality}
and is often measured in the Representational Alignment
\citep{sucholutsky2023representational_alignment}.
The Platonic Representation Hypothesis \citep{pmlr_v235_platonic_representations},
the Causal World Model Theorem \citep{richens2024casual_world_models},
metaphysical accounts Scientific Realism \citep{putnam_3_types_scientific_realism},
the Natural Latents Theory (the Natural Abstractions Hypothesis)
\citep{Wentworth_NAH, Lawrence2023_NaturalAbstractions, Wentworth_2024_natural_latents}
and the theory of Natural Kinds \citep{khalidi2023natural_kinds}
all argue for Universality of Learned Concepts in some form.

If Universality of Learned Concepts is true, then we should expect
that the problem of non-shared concepts becomes less significant
for interpretability with scale.\footnote{
    Though the level of scale that would be useful for interpretability
    is not yet clear.
}

\subsection{Universality of Learnable Concepts}

Currently, the concepts that neural networks use and learn are different from
those that humans use.
If this difference remains for the medium-term future,\footnote{
or for example, if the scale required for Universality of Learned Concepts
is far from our current scale
}
then we might be interested in whether humans can understand AI concepts that
they do not a priori use and understand.

Under our Universality framing here,
we can think of the Principle of Explanatory Optimism saying that
``even if humans and machines do not share all concepts, AI concepts are understandable
by humans \emph{given the right explanation}''.
That is, though not all concepts are \emph{human-understood}, they are \emph{human-understandable}.
The Universality of Learnable Concepts is a weaker notion than the Universality of Learned Concepts
and is directly entailed by the Universality of Learned Concepts.

\end{document}